\documentclass{acm_proc_article-sp}

\usepackage{url}
\usepackage{xcolor}
\usepackage{enumitem}

\newcommand{\vct}[1]{\ensuremath{\boldsymbol{#1}}}
\newcommand{\mat}[1]{\ensuremath{\mathtt{#1}}}
\newcommand{\set}[1]{\ensuremath{\mathcal{#1}}}
\newcommand{\con}[1]{\ensuremath{\mathsf{#1}}}
\newcommand{\T}{\ensuremath{^\top}}

\newcommand{\mean}{\operatornamewithlimits{mean}}

\newcommand{\ie}{\emph{i.e.}}
\newcommand{\eg}{\emph{e.g.}}

\makeatletter
\newcommand{\eatNext}{\@ifnextchar.\@gobble{}}
\makeatother

\newfont{\mycrnotice}{ptmr8t at 7pt}
\newfont{\myconfname}{ptmri8t at 7pt}
\let\crnotice\mycrnotice%
\let\confname\myconfname%

\permission{Preprint of the work published at AISec 2013. Please cite as: B. Biggio, I. Pillai, S. R. Bul\`o, D. Ariu, M. Pelillo, and F. Roli. Is data clustering in adversarial settings secure? In Proceedings of the 2013 ACM Workshop on Artificial Intelligence and Security, AISec '13, pages 87-98, New York, NY, USA, 2013. ACM.}

\begin{document}

\title{Is Data Clustering in Adversarial Settings Secure?}

\numberofauthors{6}
\author{
\alignauthor
Battista Biggio\\
       \affaddr{Universit\`a di Cagliari}\\
       \affaddr{Piazza d'Armi}\\
       \affaddr{09123, Cagliari, Italy}\\
       \email{battista.biggio@diee.unica.it}
\alignauthor
Ignazio Pillai\\
       \affaddr{Universit\`a di Cagliari}\\
       \affaddr{Piazza d'Armi}\\
       \affaddr{09123, Cagliari, Italy}\\
       \email{pillai@diee.unica.it}
\alignauthor Samuel Rota Bul\`o\\\
       \affaddr{FBK-irst}\\
       \affaddr{Via Sommarive, 18}\\
       \affaddr{38123, Trento, Italy}\\
       \email{rotabulo@fbk.eu}
\and  
\alignauthor Davide Ariu\\\
       \affaddr{Universit\`a di Cagliari}\\
       \affaddr{Piazza d'Armi}\\
       \affaddr{09123, Cagliari, Italy}\\
       \email{davide.ariu@diee.unica.it}
\alignauthor Marcello Pelillo\\
       \affaddr{Universit\`a Ca' Foscari di Venezia}\\
       \affaddr{Via Torino, 155}\\
       \affaddr{30172 Venezia-Mestre}\\
       \email{pelillo@dais.unive.it}
\alignauthor Fabio Roli\\
       \affaddr{Universit\`a di Cagliari}\\
       \affaddr{Piazza d'Armi}\\
       \affaddr{09123, Cagliari, Italy}\\
       \email{roli@diee.unica.it}
}

\maketitle

\begin{abstract} 
Clustering algorithms have been increasingly adopted in security applications to spot dangerous or illicit activities.
However, they have not been originally devised to deal with deliberate attack attempts that may aim to subvert the clustering process itself. Whether clustering can be safely adopted in such settings remains thus questionable.
In this work we propose a general framework that allows one to identify potential attacks against clustering algorithms, and to evaluate their impact, by making specific assumptions on the adversary's goal, knowledge of the attacked system, and capabilities of manipulating the input data. We show that an attacker may significantly poison the whole clustering process by adding a relatively small percentage of attack samples to the input data, and that some attack samples may be obfuscated to be hidden within some existing clusters. 
We present a case study on single-linkage hierarchical clustering, and report experiments on clustering of malware samples and handwritten digits.
\end{abstract}


\vspace{-4pt}



\section{Introduction}
\label{sect:introduction}

Clustering algorithms are nowadays a fundamental tool for the data analysts as they allow them to make inference and gain insights on large sets of unlabeled data. Applications of clustering span across a large number of different domains, such as market segmentation \cite{haider12,Punj83}, classification of web pages \cite{Castillo2011}, and image segmentation \cite{Forsyth2011}.
In the specific domain of computer security, clustering algorithms have been recently exploited to solve plenty of different problems, \eg, spotting fast-flux domains in DNS traffic \cite{PerdisciTDSC12}, gaining useful insights on tools and sources of attacks against Internet websites~\cite{EURECOM+2128},
detecting repackaged  Android applications \cite{HannaDIMVA2012} and (Android) mobile malware \cite{BurgueraSMSP2011}, and even automatically generating signatures for anti-virus software to enable detection of HTTP-based malware \cite{PerdisciComNet2013}.

{
  \setlength{\parskip}{3pt}
  \setlength{\parsep}{3pt}
In many of the aforementioned scenarios, a large amount of data is often collected \emph{in the wild}, in an unsupervised manner. For instance, malware samples are often collected from the Internet, by means of honeypots, \ie, machines that purposely expose known vulnerabilities to be infected by malware~\cite{Spitzner02}, or other ad hoc services, like VirusTotal.\footnote{\url{http://virustotal.com}} 
Given that these scenarios are intrinsically \emph{adversarial}, it may thus be possible for an attacker to inject carefully crafted samples into the collected data in order to subvert the clustering process, and make the inferred knowledge useless. This raises the issue of evaluating the \emph{security} of clustering algorithms against carefully designed attacks, and proposing suitable countermeasures, when required. 
It is worth noting that results from the literature of \emph{clustering stability}~\cite{Lux10} can not be directly exploited to this end, since the noise induced by adversarial manipulations is not generally \emph{stochastic} but specifically targeted against the clustering algorithm.

The problem of learning in adversarial environments has recently gained increasing popularity, and relevant research has been done especially in the area of supervised learning algorithms for classification \cite{biggio13-tkde,brueckner12,huang11,barreno-ASIACCS06}, and regression \cite{grosshans13}. On the other hand, to the best of our knowledge only few works have implicitly addressed the issue of security evaluation related to the application of clustering algorithms in adversarial settings through the definition of suitable attacks, while we are not aware of any work that proposes specific countermeasures to attacks against clustering algorithms.

The problem of devising specific attacks to subvert the clustering process was first brought to light by \textit{Dutrisac and Skillicorn} \cite{skillicorn08,skillicorn09}. They pointed out that some points can be easily \emph{hidden} within an existing cluster by forming a \emph{fringe} cluster, \ie, by placing such points sufficiently close the border of the existing cluster. They further devised an attack that consists of adding points in between two clusters to merge them, based on the notion of \emph{bridging}. Despite this pioneering attempts, a framework for the systematic security evaluation of clustering algorithms in adversarial settings is still missing, as well as a more general theory that takes into account the presence of the adversary to develop more \emph{secure} clustering algorithms. 

In this work we aim to take a first step to fill in this gap, by proposing a framework for the security evaluation of clustering algorithms, which allows us to consider several potential attack scenarios, and to devise the corresponding attacks, in a more systematic manner. Our framework, inspired from previous work on the security evaluation of supervised learning algorithms~\cite{biggio13-tkde,huang11,barreno-ASIACCS06}, is grounded on a model of the attacker that allows one to make specific assumptions on the adversary's goal, knowledge of the attacked system, and capability of manipulating the input data, and to subsequently formalize a corresponding \emph{optimal} attack strategy.
This work is thus explicitly intended to provide a cornerstone for the development of an \emph{adversarial clustering} theory, that should in turn foster research in this area. 

The proposed framework for security evaluation is presented in Sect.~\ref{sect:attacking-clustering}. In Sect.~\ref{sect:attack} we derive worst-case attacks in which the attacker has perfect knowledge of the attacked system. In particular, we formalize the notion of (worst-case) \emph{poisoning} and \emph{obfuscation} attacks against a clustering algorithm, respectively in Sects.~\ref{sect:poisoning-attacks} and \ref{sect:obfuscation-attacks}. In the former case, the adversary aims at maximally compromising the clustering output by injecting a number of carefully designed attack samples, whereas in the latter one, she tries to \emph{hide} some attack samples into an existing cluster by manipulating their feature values, without significantly altering the clustering output on the rest of the data.
As a case study, we evaluate the security of the single-linkage hierarchical clustering against poisoning and obfuscation attacks, in Sect.~\ref{sect:attacking-hierarchical}. The underlying reason is simply that the single-linkage hierarchical clustering has been widely used in security-related applications~\cite{BayerNDSS09,HannaDIMVA2012,PerdisciComNet2013,PerdisciTDSC12}. 
To cope with the computational problem of deriving an optimal attack, in Sects.~\ref{sect:poisoning-hierarchical} and~\ref{sect:obfuscation-hierarchical} we propose heuristic approaches that serve well our purposes. Finally, in Sect.~\ref{sect:experiments} we conduct synthetic and real-world experiments that demonstrate the effectiveness of the proposed attacks, and subsequently discuss limitations and future extensions of our work in Sect.~\ref{sect:conclusions}.

}

\section{Attacking Clustering}
\label{sect:attacking-clustering}

In this section we present our framework to analyze the security of clustering approaches from an adversarial pattern recognition perspective. 
It is grounded on a model of the adversary that can be exploited to identify and devise attacks against 
clustering algorithms. Our framework is inspired by a previous work focused on attacking (supervised) machine learning algorithms \cite{biggio13-tkde}, and it relies on an attack taxonomy similar to the one proposed in \cite{huang11,barreno-ASIACCS06}. 
As in \cite{biggio13-tkde}, the adversary's model entails the definition of the adversary's goal, knowledge of the attacked system, and capability of manipulating the input data, according to well-defined guidelines.

{
  \setlength{\parskip}{3pt}
  \setlength{\parsep}{3pt}
Before moving into the details of our framework, we introduce some notation.
Clustering is the problem of organizing a set of data points into groups referred to as \emph{clusters} in a way that some criteria is satisfied. 
A clustering algorithm can thus be formalized in terms of a function $f$ mapping a given dataset $\mathcal D=\{\vct x_i\}_{i=1}^{\con n}$ to 
a clustering result $\set C=f(\set D)$.
We do not specify the mathematical structure of $\set C$ at this point of our discussion because there exist different types of clustering requiring different representations, while our model applies to any of them. Indeed, $\set C$ might be a hard or soft partition of $\set D$ delivered by partitional clusterings algorithms such as k-means, fuzzy c-means or normalized cuts, or it could be a more general family of subsets of $\set D$ such as the one delivered by the dominant sets clustering algorithm \cite{PavPel07}, or it can even be a parametrized hierarchy of subsets (\eg{}, linkage-type clustering algorithms). 
}

\subsection{Adversary's goal}
\label{sect:adversary-goal}

Similarly to \cite{biggio13-tkde,huang11,barreno-ASIACCS06}, the adversary's goal can be defined according to the {attack specificity}, and the {security violation} pursued by the adversary. 
The attack specificity can be \emph{targeted}, if it affects solely the clustering of a given subset of samples; or \emph{indiscriminate}, if it potentially affects the clustering of any sample.
Security violations can instead affect the \emph{integrity} or the \emph{availability} of a system, or the \emph{privacy} of its users.

{
  \setlength{\parskip}{3pt}
  \setlength{\parsep}{3pt}

\emph{Integrity violations} amount to performing some malicious activity without significantly compromising the normal system operation. In the supervised learning setting \cite{huang11,barreno-ASIACCS06}, they are defined as attacks aiming at camouflaging some malicious samples (\eg, spam emails) to evade detection, without affecting the classification of legitimate samples. In the unsupervised setting, however, this definition can not be generally applied since the notion of malicious or legitimate class is not generally available.
Therefore, we regard integrity violations as attacks aiming at deflecting the grouping for specific samples, while limiting the changes to the original clustering. For instance, an attacker may obfuscate some samples to hide them in a different cluster, without excessively altering the initial clusters.

\emph{Availability violations} aim to compromise the functionality of the system by causing a denial of service. 
In the supervised setting, this translates into causing the largest possible classification error \cite{huang11,biggio13-tkde,biggio12-icml}.
According to the same rationale, in the unsupervised setting we can consider attacks that significantly affect
the clustering process by worsening its result as much as possible.

Finally, \emph{privacy violations} may allow the adversary to obtain information about the system's users from the clustered data by reverse-engineering the clustering process.    
}

\subsection{Adversary's knowledge}
\label{sect:adversary-knowledge}

The adversary can have different degrees of knowledge of the attacked system. They can be defined by making specific assumptions on the points $(k.i)$-$(k.iv)$ described below.

{
  \setlength{\parskip}{3pt}
  \setlength{\parsep}{3pt}

\textbf{$(k.i)$ Knowledge of the data $\set D$}: 
The adversary might know the data $\set D$ or only a portion of it. More realistically, she may not know $\set D$ exactly, but she may be able to obtain a \emph{surrogate} dataset sampled from the same distribution as $\set D$. 
In practice, this can be obtained by collecting samples from the same source from which samples in $\set D$ were collected; \eg, honeypots for malware samples~\cite{Spitzner02}.

\textbf{$(k.ii)$ Knowledge of the feature space}:
The adversary could know  how features are extracted from each sample. Similarly to the previous case, she may know how to compute the whole feature set, or only a subset of the features.

\textbf{$(k.iii)$ Knowledge of the algorithm}: 
The adversary's could be aware of the targeted clustering algorithm and how it organizes the data into clusters; \eg, the criterion used to determine the cluster set from a \emph{hierarchy} in hierarchical clustering.

\textbf{$(k.iv)$ Knowledge of the algorithm's parameters}: 
The attacker may even know how the parameters of the clustering algorithm have been initialized (if any).

\textbf{Perfect knowledge.} ~~The worst-case scenario in which the attacker has full knowledge of the attacked system, is usually referred to as \emph{perfect knowledge} case \cite{biggio13-tkde,biggio12-icml,kloft10,brueckner12,huang11,barreno-ASIACCS06}.
In our case, this amounts to knowing: $(k.i)$ the data, $(k.ii)$ the feature representation, $(k.iii)$ the clustering algorithm, and $(k.iv)$ its initialization (if any).

\subsection{Adversary's capability}
\label{sect:adversary-capability}

The adversary's capability defines how and to what extent the attacker can control the clustering process.
In the supervised setting \cite{huang11,biggio13-tkde}, the attacker can exercise a \emph{causative} or \emph{exploratory} influence, depending on whether she can control training and test data, or only test data.
In the case of clustering, however, there is not a test phase in which some data has to be classified.
Accordingly, the adversary may only exercise a \emph{causative} influence by manipulating part of the data to be clustered.\footnote{One may however think of an exploratory attack to a clustering algorithm as an attack in which the adversary aims to gain information on the clustering algorithm itself, although she may not necessarily manipulate any data to this end.} This is often the case, though, since this data is typically collected in an unsupervised manner.

{
  \setlength{\parskip}{3pt}
  \setlength{\parsep}{3pt}

We thus consider a scenario in which the attacker can add a maximum number of (potentially manipulated) samples to the dataset $\set D$. This is realistic in several practical cases, \eg, in the case of malware collected through \emph{honeypots}~\cite{Spitzner02}, where the adversary may easily send (few) samples without having access to the rest of the data. This amounts to controlling a (small) percentage of the input data.
An additional constraint may be given in terms of a maximum amount of modifications that can be done to the attack samples. In fact, to preserve their malicious functionality, malicious samples like spam emails or malware code may not be manipulated in an unconstrained manner. Such a constraint can be encoded by a suitable distance measure between the original, non-manipulated attack samples and the manipulated ones, as in \cite{biggio13-tkde,kolcz09,huang11,barreno-ASIACCS06}.

}

\subsection{Attack strategy}

Once the adversary's goal, knowledge and capabilities have been defined, one can determine an \emph{optimal} attack strategy that specifies how to manipulate the data to meet the adversary's goal, under the restriction given by the adversary's knowledge and capabilities. 
In formal terms, we denote by $\Theta$ the \emph{knowledge space} of the adversary. Elements of $\Theta$  hold information about the dataset $\set D$, the clustering algorithm $f$, and its parametrization, according to $(k.i)$-$k(.iv)$.
To model the degree of knowledge of the adversary we consider a probability distribution $\mu$ over $\Theta$. The entropy of $\mu$ indicates the level of uncertainty of the attacker. For example, if we consider a perfect-knowledge scenario like the one addressed in the next section, we have that $\mu$ is a Dirac measure peaked on an element $\theta_0\in\Theta$ (with null entropy), where $\theta_0=(\set D,f,\cdots)$ holds the information about the dataset, the algorithm and any other of the informations listed in Sect.\ref{sect:adversary-knowledge}.
Further, we assume that the adversary is given a set of attack samples $\set A$ that can be manipulated before being added to the original set $\set D$. 
We model with the function $\Omega(\set A)$ the family of sample sets that the attacker can generate according to her capability as a function of the set of initial attack samples $\set A$. The set $\set A$ can be empty, if the attack samples are not required to fulfill any constraint on their malicious functionality, \ie, they can be generated from scratch (as we will see in the case of \emph{poisoning} attacks). 
Finally, the adversary's goal given the knowledge $\theta\in\Theta$ is expressed in terms of an objective function $g(\set A';\theta)\in\mathbb R$ that evaluates how close the modified data set integrating the (potentially manipulated) attack samples $\set A'$ is to the adversary's goal.
In summary, the attack strategy boils down to finding a solution to the following optimization problem:
\begin{equation}
	\displaystyle
	\begin{array}{rl}
		\text{maximize}&\mathbb E_{\theta\sim\mu}[g(\set A';\theta)]\\
		\text{s.t.}&\set A'\in\Omega(\set A)\,.
	\end{array}
	\label{eq:optim}
\end{equation}
where $\mathbb E_{\theta\sim\mu}[\cdot]$ denotes the expectation with respect to $\theta$ being sampled according to the distribution $\mu$.

\section{Perfect knowledge attacks}
\label{sect:attack}

In this section we provide examples of worst-case integrity and availability security violations in which the attacker has perfect knowledge of the system, as described in Sect.~\ref{sect:adversary-knowledge}. We respectively refer to them as \emph{poisoning} and \emph{obfuscation} attacks.
Since the attacker has no uncertainty about the system, we set $\mu=\delta_{\{\theta_0\}}$, where $\delta$ is the Dirac measure and $\theta_0$ represents exact knowledge of the system.
The expectation in \eqref{eq:optim} thus yields $g(\set A';\theta_0)$.

}

\subsection{Poisoning attacks}
\label{sect:poisoning-attacks}

Similarly to poisoning attacks against supervised learning algorithms \cite{biggio12-icml,kloft10},
we define poisoning attacks against clustering algorithms as attacks in which the data is tainted to maximally worsen the clustering result. The adversary's goal thus amounts to violating the system's \emph{availability} by \emph{indiscriminately} altering the clustering output on any data point. 
To this end, the adversary may aim at maximizing a given distance measure between the clustering $\mathcal C$
obtained from the original data $\set D$ (in the absence of attack) and the clustering $\set C'=f_{\set D}(\set D')$ obtained by running the clustering algorithm on the contaminated data $\set D'$, and restricting the result to the samples in $\set D$, \ie{}, $f_{\set D}=\pi_{\set D}\circ f$ where $\pi_{\set D}$ is a projection operator that restricts the clustering output to the data samples in $\set D$.
We regard the tainted data $\set D'$ as the union of the original dataset $\set D$ with the attack samples in $\set A'$, \ie{}, $\set D'=\set D\cup\set A'$.
The goal can thus be written as $g(\set A';\theta_0)=d_\text{c}(\set C,f_{\set D}(\set D\cup\set A'))$, where $d_\text{c}$ is the chosen distance measure between clusterings. 
For instance, if $f$ is a partitional clustering algorithm, any clustering result can be represented in terms of a matrix $\mat Y\in\mathbb R^{\con n\times\con k}$, each $(i,k)^{\rm th}$ component being the probability that the $i^{\rm th}$ sample is assigned to the $k^{\rm th}$ cluster. Under this setting, a possible distance measure between clusterings is given by: 
\begin{equation}
	d_\text{c}(\mat Y,\mat Y')=\Vert \mat Y\mat Y\T-\mat Y'{\mat Y'}\T\Vert_F\,,
	\label{eq:distance}
\end{equation}
where $\Vert\cdot\Vert_F$ is the Frobenius norm. The components of the matrix $\mat Y\mat Y\T$ represent the probability of two samples to belong to the same cluster. When $\mat Y$ is binary, thus encoding hard clustering assignments, this distance counts the number of times two samples have been clustered together in one clustering and not in the other, or vice versa. In general, depending on the nature of the clustering result, other ad-hoc distance measures can be adopted.

{
  \setlength{\parskip}{3pt}
  \setlength{\parsep}{3pt}

As mentioned in Sect.~\ref{sect:adversary-capability}, we assume that the attacker can inject a maximum of $\con m$ data points into the original data $\mathcal D$, \ie{} $|\set A'|\leq \con m$. This realistically limits the adversary to manipulate only a given, potentially small fraction of the dataset.
Clearly, the value of $\con m$ will be considered as a parameter in our evaluation to investigate the robustness of the given clustering algorithm against an increasing control of the adversary over the data.
We further define a box constraint on the feature values $\vct x_{\rm lb} \leq \vct x \leq \mathbf x_{\rm ub}$, to restrict the attack points to lie in some fixed interval (\eg, the smallest box that includes all the data points). Hence, we define the function $\Omega$ encoding the adversary's capabilities as follows:
\[
	\Omega_\text{p}=\left\{ \{\vct a'_i\}_{i=1}^{\con m}\subset \mathbb R^{\con d}\,:\,\vct x_\text{lb}\leq \vct a'_i\leq\vct x_\text{ub}\text{ for }i=1,\cdots,\con m \right\}\,.
\]
Note that  $\Omega$ depends on a set of target samples $\set A$ in \eqref{eq:optim}, but since $\set A$ is empty in this case, we write $\Omega_\text{p}$ instead of $\Omega(\emptyset)$. The reason is simply that, in the case of a poisoning attack, the attacker aims to find a set of attack samples that do not have to carry out any specific malicious activity besides worsening the clustering process.

In summary, the optimal attack strategy  under the aforementioned hypothesis amounts to solving the following optimization problem derived from \eqref{eq:optim}:
\begin{equation}
	\displaystyle
	\begin{array}{rl}
		\text{maximize}&d_\text{c}(\set C,f_{\set D}(\set D\cup\set A'))\\
		\text{s.t.}&\set A'\in\Omega_\text{p}\,.
	\end{array}
	\label{eq:optim_pois}
\end{equation}
}

\subsection{Obfuscation attacks}
\label{sect:obfuscation-attacks}
Obfuscation attacks are violations of the system \emph{integrity} through \emph{targeted} attacks.
The adversary's goal here is to \emph{hide} a given set of initial attack samples $\set A$ within some existing clusters by \emph{obfuscating} their content, possibly without altering the clustering results for the other samples.
We denote by $\set C^t$ the target clustering involving samples in $\set D\cup\set A'$ the attacker is aiming to, being $\set A^{\prime}$ the set of obfuscated attack samples. With the intent to preserve the clustering result $\set C$ on the original data samples, we impose that $\pi_{\set D}(\set C^t)=\set C$, while the cluster assignments for the samples in $\set A'$ are freely determined by the attacker. As opposed to the poisoning attack, here the attacker is interested in pushing the final clustering towards the target clustering and therefore her intention is to minimize the distance between $\set C^t$ and $\set C'=f(\set D\cup\set A')$. Accordingly, the goal function $g$ in this case is defined as $g(\set A';\theta_0)=-d(\set C^t,f(\set D\cup \set A'))$.

{
  \setlength{\parskip}{3pt}
  \setlength{\parsep}{3pt}

As for the adversary's capability, we assume that the attacker can perturb the target samples in $\set A$ to some maximum extent. We model this by imposing that $d_\text{s}(\set A,\set A')\leq d_\text{max}$, where $d_\text{s}$ is a measure of divergence between the two sets of samples $\set A$ and $\set A'$ and $d_\text{max}$ is a nonnegative real scalar. Consequently, the function $\Omega$ representing the attacker's capacity is given by
\[
	\Omega_o(\set A)=\left\{ \{\vct a'_i\}_{i=1}^{|\set A|}\,:\, d_\text{s}(\set A,\set A')\leq d_\text{max}\right\}\,.
\]
The distance $d_\text{s}$ can be defined in different ways. For instance, in the next section we define $d_\text{s}(\set A,\set A')$ as the largest Euclidean distance among corresponding elements in $A$ and $A'$, \ie{},
\begin{equation}
	d_\text{s}(\set A,\set A')=\max_{i=1,\ldots, \con m} \Vert \vct a_i-\vct a_i'\Vert_2
	\label{eq:d_s}
\end{equation}
where we assume $\set A=\{\vct a_i\}_{i=1}^{\con m}$ and $\set A'=\{\vct a'_i\}_{i=1}^{\con m}$.
This choice allows us to bound the divergence between the original target samples in $\set A$ and the manipulated ones, as typically done in adversarial learning \cite{kolcz09,huang11,brueckner12,biggio13-tkde}.

In summary, the attack strategy in the case of obfuscation attacks can be obtained as the solution of the following optimization program derived from \eqref{eq:optim}:
\begin{equation}
	\displaystyle
	\begin{array}{rl}
		\text{minimize}&d_\text{c}(\set C^t,f(\set D\cup\set A'))\\
		\text{s.t.}&\set A'\in\Omega_\text{o}(\set A)\,.
	\end{array}
	\label{eq:optim_obfu}
\end{equation}
}

\section{A case study on single-linkage hierarchical clustering}
\label{sect:attacking-hierarchical}

In this section we solve a particular instance of the optimization problems \eqref{eq:optim_pois} and \eqref{eq:optim_obfu}, corresponding respectively to the poisoning and obfuscation attacks described in Sects.~\ref{sect:poisoning-attacks} and \ref{sect:obfuscation-attacks}, against the single-linkage hierarchical clustering. The motivation behind this specific choice of clustering algorithm is that, as mentioned in Sect.~\ref{sect:introduction}, it has been frequently exploited in security-sensitive tasks~\cite{BayerNDSS09,HannaDIMVA2012,PerdisciComNet2013,PerdisciTDSC12}.

Single-linkage hierarchical clustering is a bottom-up algorithm that produces a \emph{hierarchy} of clusterings, as any other hierarchical \emph{agglomerative} clustering algorithm \cite{JainACD1988}. The hierarchy is represented by a \emph{dendrogram}, \ie, a tree-like data structure showing the sequence of cluster fusion together with the distance at which each fusion took place.
To obtain a given partitioning of the data into clusters, the dendrogram has to be \emph{cut} at a certain height. The leaves that form a connected sub-graph after the cut are considered part of the same cluster.
Depending on the chosen distance between clusters (\emph{linkage} criterion), different variants of hierarchical clustering can be defined.
In the \emph{single-linkage} variant, the distance between any two clusters $\set C_{1}, \set C_{2}$ is defined as the minimum Euclidean distance between all pairs of samples in $\set C_1\times\set C_2$. 

{
  \setlength{\parskip}{3pt}
  \setlength{\parsep}{3pt}

For both poisoning and obfuscation attacks, we will model the clustering output as a binary matrix $\mat Y\in\{0,1\}^{\con n\times\con k}$, indicating the sample-to-cluster assignments (see Sect.~\ref{sect:poisoning-attacks}).
Consequently, we can make use of the distance measure $d_{c}$ between clusterings defined in Eq.~\eqref{eq:distance}. However, to obtain a given set of clusters from the dendrogram obtained by the single-linkage clustering algorithm, we will have to specify an appropriate \emph{cut} criterion.

}

\subsection{Poisoning attacks}
\label{sect:poisoning-hierarchical}

\begin{figure*}[t]
\centering
\includegraphics[width=0.32\textwidth]{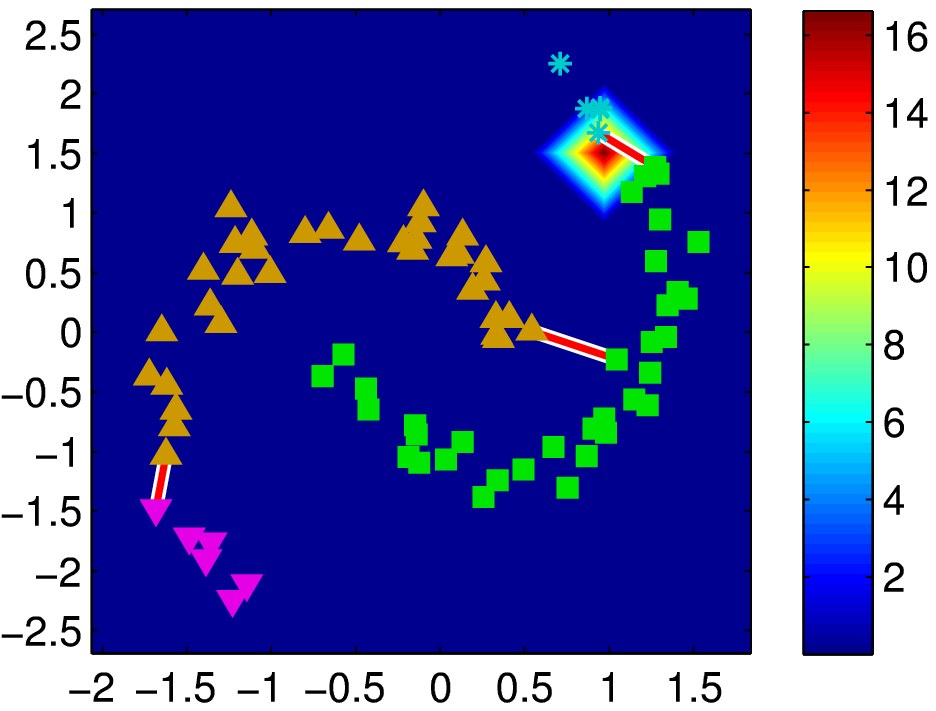}
\includegraphics[width=0.32\textwidth]{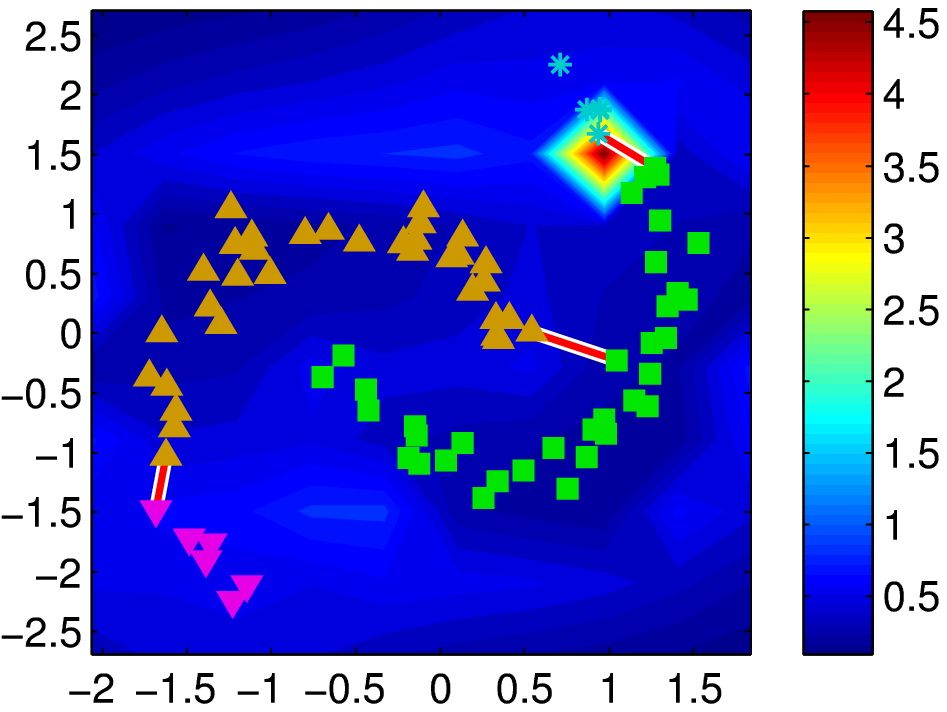}
\includegraphics[width=0.32\textwidth]{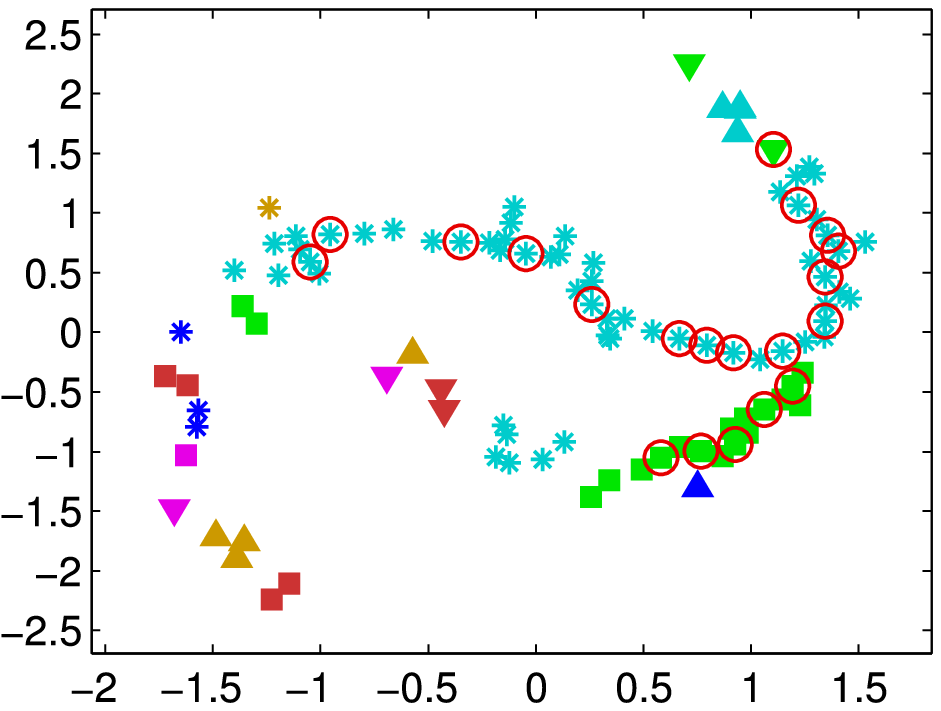}
\vspace{-10pt}
\caption{Poisoning single-linkage hierarchical clustering. In each plot, samples belonging to different clusters are represented with different markers and colors.
The left and middle plot show the initial partitioning of the given 100 data points into $k=4$ clusters.
The objective function of Eq.~\ref{eq:optim_pois} (shown in colors) for our greedy attack ($|\mathcal A^{\prime}|=1$) is respectively computed with hard (left plot) and soft assignments (middle plot), \ie, with binary $\mat Y$ and posterior estimates. The $k-1=3$ \emph{bridges} obtained from the dendrogram are highlighted with red lines. The rightmost plot shows how the partitioning changes after $\con m=20$ attack samples (highlighted with red circles) have been greedily added.}
\label{fig:example}
\vspace{-10pt}
\end{figure*}

For poisoning attacks against single-linkage hierarchical clustering, we aim to solve the optimization problem given by Eq.~\eqref{eq:optim_pois}. 
As already mentioned, since the clustering is expressed in terms of a hierarchy, we 
have to determine a suitable dendrogram cut in order to model the clustering output as a binary matrix $\mat Y$.
In this case, we assume that the clustering algorithm selects the cut, \ie, the number of clusters, that achieves the minimum distance between the clustering obtained in the absence of attack $\set C$ and the one induced by the cut, \ie, $\min d_c(\set C,f_{\set D}(\set D\cup\set A'))$. 
Although this may not be a realistic \emph{cut} criterion, as the ideal clustering $\set C$ is not known to the clustering algorithm, this worst-case choice for the adversary gives us the minimum performance degradation incurred by the clustering algorithm under attack. 

{
  \setlength{\parskip}{3pt}
  \setlength{\parsep}{3pt}

Let us now discuss how Problem~\eqref{eq:optim_pois} can be solved. First, note that it is not possible to predict analytically how the clustering output $\mat Y^{\prime}$ changes as the set of attack samples $\set A^{\prime}$ is altered, since hierarchical clustering does not have a tractable, underlying analytical interpretation.\footnote{In general, even if the clustering algorithm has a clearer mathematical formulation, it is not guaranteed that a good analytical prediction can be found. For instance, though k-means clustering is well-understood mathematically, its variability to different initializations makes it almost impossible to reliably predict how its output may change due to data perturbation.}
One possible answer consists in a stochastic exploration of the solution space (\eg{} by simulated annealing). 
This is essentially done by perturbing the input data $\set A^{\prime}$ a number of times, and evaluating the corresponding values of the objective function by running the clustering algorithm (as a black box) on $\set D \cup \set A^{\prime}$. The set $\set A^{\prime}$ that provides the highest objective value is eventually retained. However, to find an optimal configuration of attack samples $\set A^{\prime}$, one should repeat this procedure a very large number of times. To reduce computational complexity, one may thus consider efficient search heuristics specifically tailored to the considered clustering algorithm.

For the above reason, we consider a greedy optimization approach where the attacker aims at finding a local maximum of the objective function by adding one attack sample at a time, \ie, $|\set A^{\prime}|=\con m=1$. In this case, we can more easily understand how the objective function changes as the inserted attack point varies, and define a suitable heuristic approach. An example is shown in the leftmost plot of Fig.~\ref{fig:example}. This plot shows that the objective function exhibits a global maximum when the attack point is added in between clusters that are sufficiently \emph{close} to each other. The reason is that, when added in such a location, the attack point operates as a \emph{bridge}, causing the two clusters to be merged in a single cluster, and the objective function to increase.

\textbf{Bridge-based heuristic search}. Based on this observation, we devised a search heuristic that considers only $k-1$ potential attack samples, being $k$ the actual number of clusters found by the single-linkage hierarchical clustering at a given dendrogram cut. In particular, we only considered the $k-1$ points lying in between the connections that have been cut to separate the $k$ given clusters from the top of the hierarchy, highlighted in our example in the leftmost plot of Fig.~\ref{fig:example}.
These connections can be directly obtained from the dendrogram, \ie, we do not have to run any post-processing algorithm on the clustering result. Thus, one is only required to evaluate the objective function $k-1$ times for selecting the best attack point. We will refer to this approach as \emph{Bridge (Best)} in Sect.~\ref{sect:experiments-poisoning}. The rightmost plot in Fig.~\ref{fig:example} shows the effect of our greedy attack after that $\con m=20$ attack points have been inserted. Note how the initial clusters are fragmented into smaller clusters that tend to contain points which originally belonged to different clusters.

\textbf{Approximating $\mat Y^{\prime}$}. To further reduce the computational complexity of our approach, \ie, to avoid re-computing the clustering and the corresponding value of the objective function $k-1$ times for each attack point, we consider another heuristic approach. 
The underlying idea is simply to select the attack sample (among the $k-1$ \emph{bridges} suggested by our bridge-based heuristic search) that lies in between the largest clusters. In particular, we assume that the attack point will effectively merge the two adjacent clusters, and thus modify $\mat Y^{\prime}$ accordingly (without re-estimating its real value by re-running the clustering algorithm).
To this end, for each point belonging to one of the two clusters, we set to $1$ ($0$) the value of $\mat Y^{\prime}$ corresponding to the first (second) cluster.
Once the estimated $\mat Y^{\prime}$ is computed, we evaluate the objective function using the estimated $\mat Y^{\prime}$, and select the attack point that maximizes its value.
We will refer to this approach as \emph{Bridge (Hard)} in Sect.~\ref{sect:experiments-poisoning}.

\textbf{Approximating $\mat Y^{\prime}$ with soft clustering assignments}. Finally, we discuss another variation to the latter discussed heuristic approach, which we will refer to as \emph{Bridge (Soft)}, in Sect.~\ref{sect:experiments-poisoning}.
The problem arises from the fact that our objective function exhibits really \emph{abrupt} variations, since it is computed on hard cluster assignments (\ie, binary matrices $\mat Y^{\prime}$). 
Accordingly, adding a single attack point at a time may not reveal \emph{connections} that can potentially merge large clusters after few attack iterations, \ie, using more than one attack sample.
To address this issue, we approximate $\mat Y^{\prime}$ with soft clustering assignments. To this end, the element $y^{\prime}_{ik}$ of $\mat Y^{\prime}$ is estimated as the posterior probability of point $\vct x_{i}$ belonging to cluster $c_{k}$, \ie, $y^{\prime}_{ik}=p(c_{k} | \vct x_{i}) =  p(\vct x_{i} | c_{k})p(c_{k})/p(\vct x_{i})$. The prior $p(c_{k})$ is estimated as the number of samples belonging to $c_{k}$ divided by the total number of samples, the likelihood $p(\vct x_{i} | c_{k})$ is estimated with a Gaussian Kernel Density Estimator (KDE) with bandwidth parameter $h$:
\begin{equation}
p(\vct x_{i} | c_{k}) = \frac{1}{|c_{k}|} \sum_{\vct x_{j} \in c_{k}} \exp \left (-\frac{||\vct x_{i}-\vct x_{j}||^{2}}{h} \right )  \enspace ,
\label{eq:kde}
\end{equation} 
and the evidence $p(\vct x_{i})$ is obtained by marginalization over the given set of clusters.

Worth noting, for too small values of $h$, the posterior estimates tend to the same value, \ie, each point is likely to be assigned to any cluster with the same probability. When $h$ is too high, instead, each point is assigned to one cluster, and the objective function thus equals that corresponding to the original hard assignments. In our experiments we simply avoid these limit cases by selecting a value of $h$ comparable to the average distance between all possible pairs of samples in the dataset, which gave reasonable results.

An example of the smoother approximation of the objective function provided by this heuristic is shown in the middle plot of Fig.~\ref{fig:example}. Besides, this technique also provides a \emph{reliable} approximation of the true objective: although its values are significantly re-scaled, the global maximum is still found in the same location. The smooth variations that characterize the approximated objective influence the choice of the best candidate attack point. In fact, attack points lying on \emph{bridges} that may potentially connect larger clusters after some attack iterations may be sometimes preferred to attack points that can directly connect smaller and closer clusters. This may lead to a larger increase in the true objective function as the number of injected attack points increases. 

}

\subsection{Obfuscation attacks}
\label{sect:obfuscation-hierarchical}

In this section we solve \eqref{eq:optim_obfu} assuming the worst-case (perfect-knowledge) scenario against the single-linkage clustering algorithm.
Recall that the attacker's goal in this case is to manipulate a given set of non-obfuscated samples $\set A$ such that they are clustered according to a desired configuration, \eg{}, together with points in an existing, given cluster, without altering significantly the initial clustering that would be obtained in the absence of manipulated  attacks.

{
  \setlength{\parskip}{3pt}
  \setlength{\parsep}{3pt}
As in the previous case, to represent the output of the clustering algorithm as a binary matrix $\mat Y$ representing clustering assignments, and thus compute $d_{c}$ as given by Eq.~\ref{eq:distance}, we have to define a suitable criterion for cutting the dendrogram.
Similarly to poisoning attacks, we define an advantageous criterion for the clustering algorithm, that gives us the lowest performance degradation incurred under this attack: we select the dendrogram cut that minimizes $d_{c}(\set C^{\star},f(\set D \cup \set A^{\prime}))$, where $\set C^{\star}$ represents the optimal clustering that would be obtained including the non-manipulated attack samples, \ie{}, $\set C^{\star}=f(\set D \cup \set A)$. The reason is that, to better contrast an obfuscation attack, the clustering algorithm should try to keep the attack points corresponding to the non-manipulated set $\set A$ into their original clusters. For instance, in the case of malware clustering, non-obfuscated malware may easily end up in a well-defined cluster, and, thus, it may be subsequently categorized in a well-behaved malware family. While the adversary tries to manipulate malware to have it clustered differently, the best solution for the clustering algorithm would be to obtain the same clusters that would be obtained in the absence of attack manipulation.

We derive a simple heuristic to get an approximate solution of \eqref{eq:optim_obfu} assuming $d_\text{s}$ to be defined as in \eqref{eq:d_s}.
We assume that, for each sample $\vct a_i\in\set A$, the attacker selects the closest sample $\vct d_i\in\set D$ belonging to the cluster to which $\vct a_i$ should belong to, according to the attacker's desired clustering $\set C^{t}$.
To meets the constraint given by $\Omega_\text{o}$ in Eq.~\ref{eq:optim_obfu}, the attacker then determines for each $\vct a_i\in\set A$ a new sample $\vct a'_i\in\set A$ along the line connecting $\vct a_i$ and $\vct d_i$ in a way not to exceed the maximum distance $d_\text{max}$ from $\vct a_i$, \ie,
$\vct a_i'=\vct a_i+\alpha (\vct d_i-\vct a_i)$, 
where $\alpha=\min(1,d_\text{max}/\Vert\vct d_i-\vct a_i\Vert_2)$.
}

\section{Experiments}
\label{sect:experiments}

We present here some experiments to evaluate the effectiveness of the poisoning and obfuscation attacks devised in Sect.~\ref{sect:attacking-hierarchical} against the single-linkage hierarchical clustering algorithm, under perfect knowledge of the attacked system.

\subsection{Experiments on poisoning attacks}
\label{sect:experiments-poisoning}

For the poisoning attack, we consider three distinct cases: a two-dimensional artificial data set, a realistic application example on malware clustering, and a task in which we aim to cluster together distinct handwritten digits.

\begin{figure*}[t]
\centering
\includegraphics[width=0.32\textwidth]{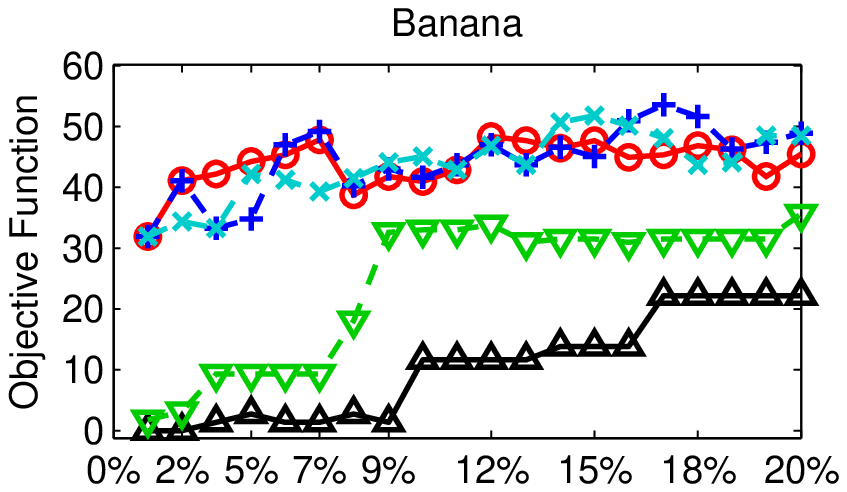}
\includegraphics[width=0.32\textwidth]{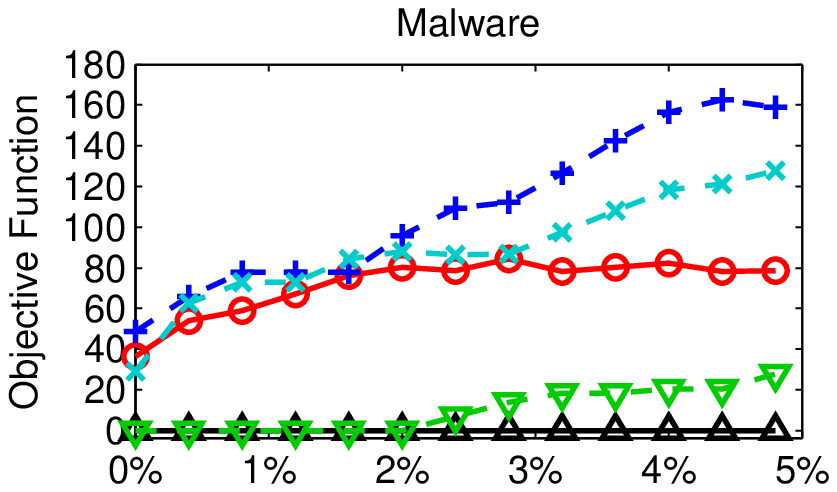}
\includegraphics[width=0.32\textwidth]{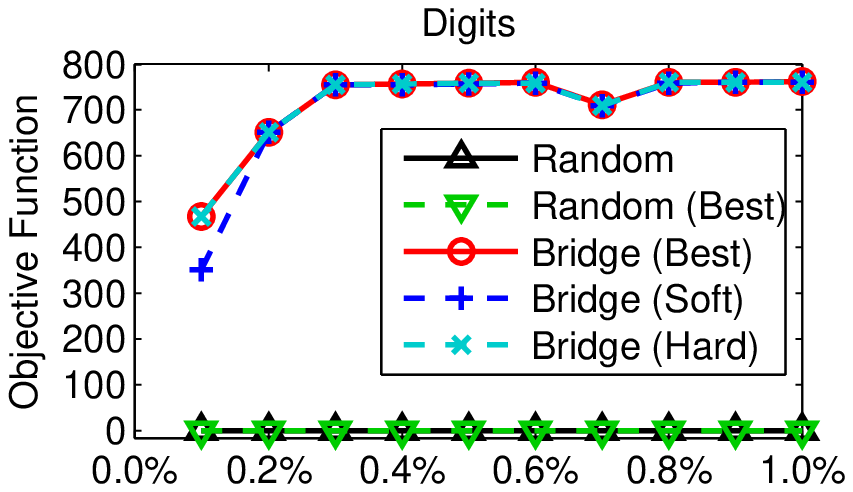}
\includegraphics[width=0.32\textwidth]{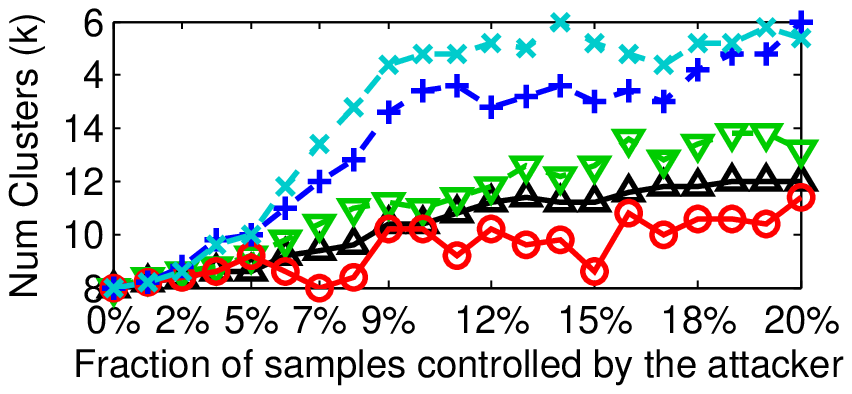}
\includegraphics[width=0.32\textwidth]{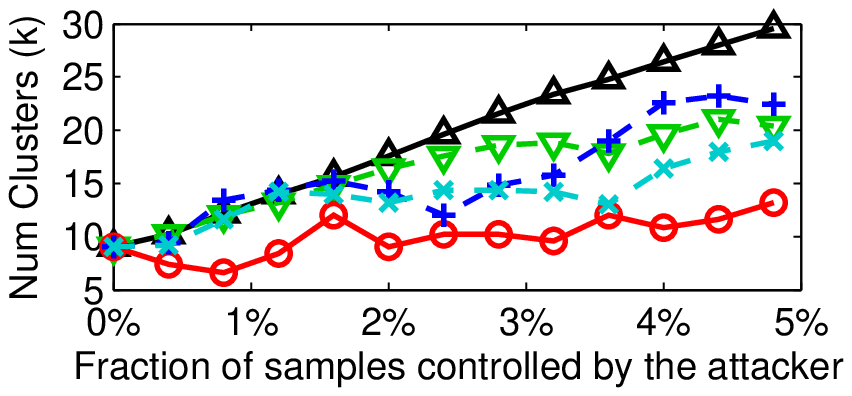}
\includegraphics[width=0.32\textwidth]{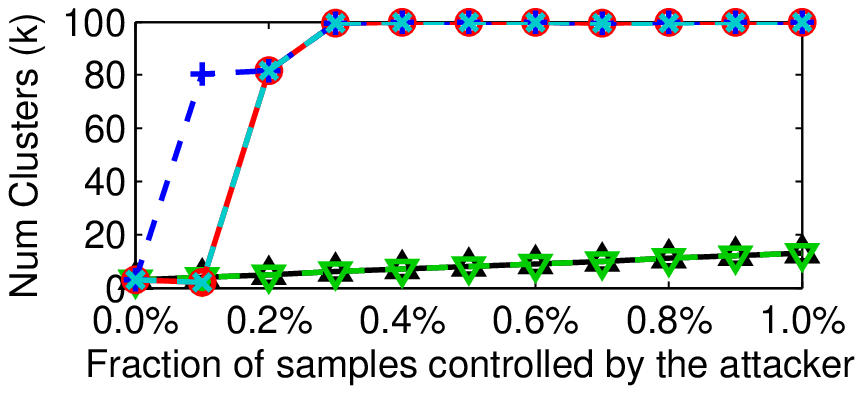}
\vspace{-10pt}
\caption{Results for the poisoning attack averaged over five runs on the Banana-shaped dataset (first column), the Malware dataset (second column), and the Digit dataset (third column). Top plots show the variation of the objective function $d_\text{c}(f(\set D),f_{\set D}(\set D\cup\set A'))$ as the fraction of samples controlled by the adversary increases. Bottom plots report the number of clusters selected after the insertion of each attack sample.}
\label{fig:results}
\end{figure*}

\begin{table*}
\label{tab:split-merge}
\centering
\begin{tabular}{l l@{ $\pm$ }rl@{ $\pm$ }r l@{ $\pm$ }rl@{ $\pm$ }r l@{ $\pm$ }rl@{ $\pm$ }r}
\hline
&\multicolumn{4}{c}{Banana (20\%)} & \multicolumn{4}{c}{Malware (5\%)} & \multicolumn{4}{c}{Digits (1\%)} \\
& \multicolumn{2}{c}{\emph{Split}} & \multicolumn{2}{c}{\emph{Merge}} & \multicolumn{2}{c}{\emph{Split}} & \multicolumn{2}{c}{\emph{Merge}} & \multicolumn{2}{c}{\emph{Split}} & \multicolumn{2}{c}{\emph{Merge}}  \\
\hline
\hline
Random        & 1.15 & 0.22 & 1.29 & 0.06         & 1.00 & 0.00 & 1.00 & 0.00 &      1.00 & 0.00 & 1.00 & 0.00   \\
Random (Best) & 1.40 & 0.34 & 1.54 & 0.30         & 1.00 & 0.00 & 1.34 & 0.39 &      1.00 & 0.00 & 1.00 & 0.00   \\
Bridge (Best) & 2.40 & 0.60 & 1.40 & 0.23         & 1.49 & 0.23 & 1.31 & 0.17 &      33.9 & 0.15 & 1.02 & 0.00   \\
Bridge (Soft) & 3.85 & 1.35 & 1.22 & 0.11         & 2.76 & 0.84 & 1.12 & 0.09 &      33.9 & 0.15 & 1.02 & 0.00   \\
Bridge (Hard) & 3.75 & 1.43 & 1.21 & 0.23         & 2.41 & 0.73 & 1.10 & 0.10 &      34.0 & 0.00 & 1.02 & 0.00   \\
\hline
\end{tabular}
\vspace{-5pt}
\caption{Split and Merge averaged values and standard deviations for the Banana-shaped dataset (at 20\% poisoning), the Malware dataset (at 5\% poisoning), and the Digit dataset (at 1\% poisoning).}
\vspace{-2pt}
\end{table*}

\subsubsection{Artificial data}
\label{sect:artificial}

We consider here the standard two-dimensional banana-shaped dataset from PRTools,\footnote{\url{http://prtools.org}} for which a particular instance is shown in Fig.~\ref{fig:example} (right and middle plot).
We fix the number of initial clusters to $k=4$, which yields our original clustering $\set C$ in the absence of attack.

{
  \setlength{\parskip}{3pt}
  \setlength{\parsep}{3pt}

We repeat the experiment five times, each time by randomly sampling 80 data points.
In each run, we add up to $\con m = 20$ attack samples, that simulates a scenario in which the adversary can control up to 20\% of the data.
As described in Sect.~\ref{sect:poisoning-hierarchical}, the attack proceeds greedily by adding one sample at a time. After adding each attack sample, we allow the clustering algorithm to change the number of clusters from a minimum of $2$ to a maximum of $50$. The criterion used to determine the number of clusters is to minimize the distance of the current partitioning with the clustering in the absence of attack, as explained in details in Sect.~\ref{sect:poisoning-hierarchical}.

We consider five attack strategies, described in the following.

\textit{Random}: the attack point is selected at random in the minimum box that encloses the data.

\textit{Random (Best)}: $k-1$ attack points are selected at random, being $k$ the actual number of clusters at a given attack iteration. Then, the objective function is evaluated for each point, and the best one is chosen.

\textit{Bridge (Best)}: The $k-1$ bridges suggested by our heuristic approach are evaluated, and the best one is chosen.

\textit{Bridge (Hard)}: The $k-1$ bridges are evaluated here by predicting the clustering output $\mat Y^{\prime}$ as discussed in Sect.~\ref{sect:poisoning-hierarchical} (\ie, assuming that the corresponding clusters will be merged), using hard clustering assignments.

\textit{Bridge (Soft)}: This is the same strategy as \emph{Bridge (Hard)}, except for the fact that we consider soft clustering assignments when modifying $\mat Y^{\prime}$. To this end, as discussed in Sect.~\ref{sect:poisoning-hierarchical}, we use a Gaussian KDE. We set the kernel bandwidth $h$ as the average distance between each possible pair of samples in the data. On average, $h \approx 2$ in each run.

It is worth remarking that \emph{Random (Best)} and \emph{Bridge (Best)} require the objective function to be evaluated $k-1$ times at each iteration to select the best candidate attack sample. This means that the clustering algorithm has to be run $k-1$ times at each step. Instead, the other methods do not require us to re-run the clustering algorithm to select the attack point. Their complexity is therefore significantly lower than the aforementioned methods.

The results averaged over the five runs are reported in Fig.~\ref{fig:results} (first column).
From the top plot one may appreciate how the methods based on the \emph{bridge}-based heuristics achieve similar values of the objective function, while clearly outperforming the \emph{random}-based methods.
Further, as reasonably expected, \emph{Random (Best)} outperforms \emph{Random} since it considers the best point over $k-1$ attempts. Nevertheless, even selecting a random attack sample, in this case, turned out to significantly affect the clustering results.

The bottom plot provides us a better understanding of how the attack effectively works. The main effect is indeed to \emph{fragment} the original clusters into a high number of smaller clusters. In particular, after the insertion of $\con m=20$ data points, \ie, when 20\% of the data is controlled by the attacker, the selected number of clusters increases from 4 to about 7-14 clusters depending on the considered method. 

To further clarify the effect of the attack on the clustering algorithm, we consider two measures referred to as \emph{Split} and \emph{Merge} in Table~\ref{tab:split-merge}, which are given as follows. Let $\set C$ and $\set C'$ be the initial and the final clustering restricted to elements in $\set D$, respectively, and let $\mat C$ be a binary matrix, each entry $\mat C_{kk'}$ indicating the co-occurrence of at least one sample in the $k$th cluster of $\set C$ and in the $k'$th cluster of $C'$. Then, the above measures are given as:
\[
{\rm Split}=\mean_i \sum_j \mat C_{ij}\,, \, {\rm Merge}=\mean_j \sum_i \mat C_{ij}\,.
\]
Intuitively, \emph{split} quantifies to what extent the initial clusters are fragmented across different final clusters, while \emph{merge} quantifies to what extent the final clusters contain samples that originally belonged to different initial clusters.

From Table~\ref{tab:split-merge}, it can be appreciated how, for the most effective attacks, \ie, \emph{Bridge (Soft)} and \emph{Bridge (Hard)}, the initial clusters are split into approximately 3.8 clusters, while the final clusters merge approximately 1.2 initial clusters, on average.
This clarifies how the proposed attack eventually compromises the initial clustering: it tends to fragment the initial clusters into smaller ones, and to merge together final clusters which originally came from different clusters.
\emph{Bridge (Best)} tends instead to induce a lower number of final clusters, \ie, the clustering algorithm tends to merge more final clusters than splitting initial ones. However, this is not the optimal choice according to the attacker's goal.

\begin{figure}[t]
\centering
\includegraphics[width=0.48\textwidth]{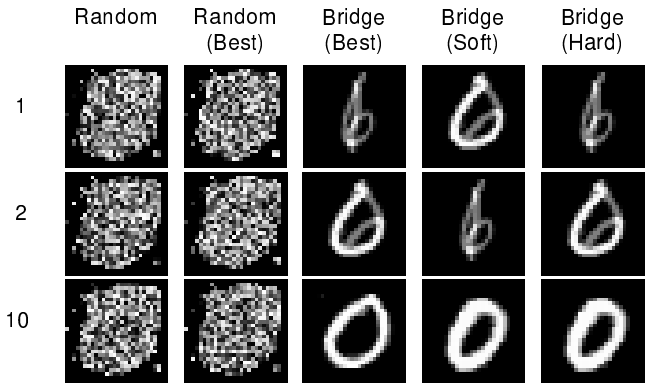}
\vspace{-15pt}
\caption{Attack samples produced by the five strategies at iterations 1, 2 and 10, for the digit data.}
\label{fig:results-mnits}
\vspace{-10pt}
\end{figure}

}

\subsubsection{Malware clustering}
\label{sect:malware-clustering}

We consider here a more realistic application example involving malware clustering, and in particular a simplified version of the algorithm for behavioral malware clustering proposed in \cite{PerdisciComNet2013}. The ultimate goal of this approach is to obtain malware clusters that can aid the automatic generation of high quality network signatures, which can be used in turn to detect botnet command-and-control (C\&C) and other malware-generated communications at the network perimeter. 
With respect to the original algorithm, we made the following simplifications:
\vspace{-5pt}
\begin{enumerate}[leftmargin=*]
		  \setlength{\itemsep}{1pt}
  \setlength{\parskip}{0pt}
  \setlength{\parsep}{0pt}
\item[(a)] we consider only the first of the two clustering steps carried out by the original system. The algorithm proposed in \cite{PerdisciComNet2013} clusters samples through two consecutive stages, named \emph{coarse-grain} and \emph{fine-grain} clustering, respectively. Here, we just focus on the \emph{coarse-grain} clustering, which is based on a set of numeric features.
\item[(b)] We consider a subset of six statistical features (out of the seven used by the original algorithm). They are: $(1)$ number of GET requests; $(2)$ number of POST requests; $(3)$ average length of the URLs; $(4)$ average number of parameters in the request; $(5)$ average amount of data sent by POST requests; and $(6)$ average length of the response. We exclude the seventh feature, \ie, the total number of HTTP requests, as it is redundant with respect to the first and the second feature. All feature values are re-scaled in $[0,1]$ as in the original work.
\item[(c)] We use the single-linkage hierarchical clustering instead of the BIRCH algorithm \cite{BIRCH}, since this modification does not significantly affect the quality of the clustering results, as the authors demonstrated in \cite{PerdisciComNet2013}.
\end{enumerate}
\vspace{-5 pt}

{
  \setlength{\parskip}{3pt}
  \setlength{\parsep}{3pt}

For the purpose of this evaluation, we use a subset of 1,000 samples taken from \textit{Dataset 1} of \cite{PerdisciComNet2013}. This dataset consists of distinct malware samples (no duplicates) collected during March 2010 from a number of different malware sources, including MWCollect~\cite{MWCollect}, Malfease~\cite{Malfease}, and commercial malware feeds. 
As in the previous setting, we repeat the experiments five times, by randomly selecting a subset of $475$ samples from the available set of $1,000$ malware data in each run.
The initial set of clusters $\set C$, as in \cite{PerdisciComNet2013}, is selected as the partitioning that minimizes the value of the Davies-Bouldin Index (DBI) \cite{HalkidiJIIS2001}, a measure that characterizes dispersion and closeness of clusters. We consider the cuts of the initial dendrogram that yield from 2 to 25 clusters, and choose the one corresponding to the minimum DBI. This yields approximately $9$ clusters in each run. 
While the attack proceeds, the clustering algorithm can choose a number of clusters ranging from $2$ to $50$.
The attacker can inject up to 25 attack samples, that amounts to controlling up to 5\% of the data.
The value of $h$ for the KDE used in \emph{Bridge (Soft)} is set as the average distance between pairs of samples, which turns out to be approximately $0.2$ in each run. 

Results are shown in Fig.~\ref{fig:results} (second column). The effect of the attack is essentially the same as in the previous experiments on the Banana-shaped data, although here there is a significant difference among the performances of the \emph{bridge}-based methods. In particular, \emph{Bridge (Soft)} gradually outperforms the other approaches as the fraction of injected samples approaches 5\%. The reason is that, as qualitatively discussed in Sect.~\ref{sect:poisoning-hierarchical}, this heuristic approach tends to bridge clusters which are too far to be bridged with a single attack point, and are thus disregarded by \emph{Bridge (Best)} and not always chosen by \emph{Bridge (Hard)}. It is also worth noting that, in this case, the Random approach is totally ineffective. In particular, no change in the objective function is observed for this method, and the number of clusters increases linearly as the attack proceeds. This means simply that the clustering algorithm produces a new cluster for each newly-injected attack point, making the attack totally ineffective. The behavior exhibited by the different attack strategies is also confirmed by the Split and Merge values reported in Table~\ref{tab:split-merge}. Here, the most effective methods, \ie, again \emph{Bridge (Soft)} and \emph{Bridge (Hard)}, split the 3 initial clusters each into 2.7 and 2.4 final clusters, on average, yielding a total number of clusters of about 20-25 clusters. Similarly to the previous experiments, \emph{Bridge (Best)} yields a lower number of final clusters, as it induces more the clustering algorithm to cluster together samples that originally belonged to different initial clusters.
}

\subsubsection{Handwritten digits}
\label{sect:mnist}

We finally repeat the experiments described in the previous sections on the MNIST handwritten digit data \cite{LeCun95}.\footnote{This dataset is publicly available in Matlab format at \url{http://cs.nyu.edu/~roweis/data.html}.}
In this dataset, each digit is size-normalized and centered, and represented as a grayscale image of $28 \times 28$ pixels. Each pixel is raster-scan ordered and its value is directly considered as a feature. The dimensionality of the feature space is thus $784$, a much higher value than that considered in the previous cases. We further normalize each feature (pixel) in $[0,1]$ by dividing its value by $255$.

{
  \setlength{\parskip}{3pt}
  \setlength{\parsep}{3pt}

We focus here on a subset of data consisting of the three digits `0', `1', and `6'.
To obtain three initial clusters, each representing one of the considered digits, we first compute the average digit for each class (\ie, the average `0', `1', and `6'), and then select $700$ samples per class, by retaining the closest samples to the corresponding average digit.
We repeat the experiments five times, each time by randomly selecting $330$ samples per digit from the corresponding set of $700$ pre-selected samples.
While the attack proceeds, the clustering algorithm can choose a number of clusters ranging from $2$ to $100$.
We assume that the attacker can inject up to $10$ attack samples, that amounts to controlling up to 1\% of the data. The value of $h$ for the KDE used in \emph{Bridge (Soft)} is set as in the previous case, based on the average distance between all pairs of samples. For this dataset, it turns out that $h\approx1$ in each run. 

Results are shown in Fig.~\ref{fig:results} (third column). With respect to the previous experiments on the Banana-shaped data, and on the Malware data, the results here are significantly different.
In particular, note how the \emph{Random} and \emph{Random (Best)} approaches are totally ineffective here.
Similarly to the previous case in malware clustering, the clustering algorithm essentially defeats the attack influence by creating a new cluster for each attack sample. The underlying reason is that, in this case, the feature space has a very high dimensionality, and, thus, sampling only $k-1$ points at random is not enough to find a suitable attack point. In other words, if an attack sample is not very well crafted, it may be easily isolated from the rest of the data.
Although increasing the dimensionality may thus seem a suitable countermeasure to protect clustering against random attacks, this drastically increases its vulnerability to well designed attack samples. Note indeed how the clustering is already significantly worsened when the adversary only controls a fraction as small as of 0.2\% of the data. In fact, the number of final clusters raises immediately to the maximum allowed number of 100. This is also clarified in Table~\ref{tab:split-merge}, where it can be appreciated how the initial clusters are fragmented into an average of 33 final clusters for the \emph{bridge}-based methods.
Note however that, in this case, the final clusters are almost \emph{pure}, \ie, the attack algorithm does not succeed in merging together samples coming from different initial clusters.

In Fig.~\ref{fig:results-mnits} we also show some of the attack samples that are produced by the five attack strategies, at different attack iterations.
The \emph{random}-based attacks clearly produce very noisy images which yield a completely ineffective attack, as already mentioned. Instead, the initial attacks considered by \emph{bridge}-based methods (at iteration 1 and 2) resemble effectively the digits corresponding to the two initial clusters that they aim to connect (`0' and `6', and `1' and`6'). Since the attack completely destroys the three initial clusters after very few attack samples have been added, at later iterations (\eg, iteration 10), the \emph{bridge}-based methods tend to enforce some connection within the cluster belonging to the `0' digit, probably trying to merge some of the final clusters together. However, since the maximum number of allowed clusters has been already reached, no further improvement is observed in the objective function.
}

\subsection{Experiments on obfuscation attacks}
\label{sect:experiments-obfuscation}

For the obfuscation attack, we present an experiment on handwritten digits, using again the MNIST digit data described in Sect.~\ref{sect:mnist}.

\subsubsection{Handwritten digits}
\label{sect:mnist-obfuscation}

We consider the same initial clusters of Sect.~\ref{sect:mnist}, consisting of 330 samples for each of the following digits: `0', `1', and `6'. As in the previous case, we average the results over five runs, each time selecting the initial 330 samples per cluster from the pre-selected sets of 700 samples per digit.
In this case, however, we consider a further initial cluster of 100 samples corresponding to the digit `3' (which are also randomly sampled from a pre-selected set of 700 samples of `3', chosen with the same criterion used in Sect.~\ref{sect:mnist} to end up in the same cluster, initially).
These represent the attack samples $\set A$ that the attacker aims to obfuscate.
We remind the reader that the attacker's goal in this case is to manipulate some samples to have them clustered according to a desired criterion, without affecting significantly the normal system operation.
In particular, we assume here that the attacker can manipulate samples corresponding to the digit `3'  in order to have them clustered together with the cluster corresponding to the digit `6', while preserving the initial clusters.
In other words, the desired clustering output for the attacker consists of three clusters: one corresponding to the `0' digit, one corresponding to the `1' digit, and the latter corresponding to the digits `6' and `3'.
These constraints can be easily encoded as a desired clustering output $\set C^{t}$ through a binary matrix $\mat Y^{t}$. This reflects exactly Problem~\ref{eq:optim_obfu}, where the attacker aims at minimizing $d_{c}( \set C^{t},f(\set D \cup \set A^{\prime}))$.

\begin{figure}[t]
\centering
\includegraphics[width=0.32\textwidth]{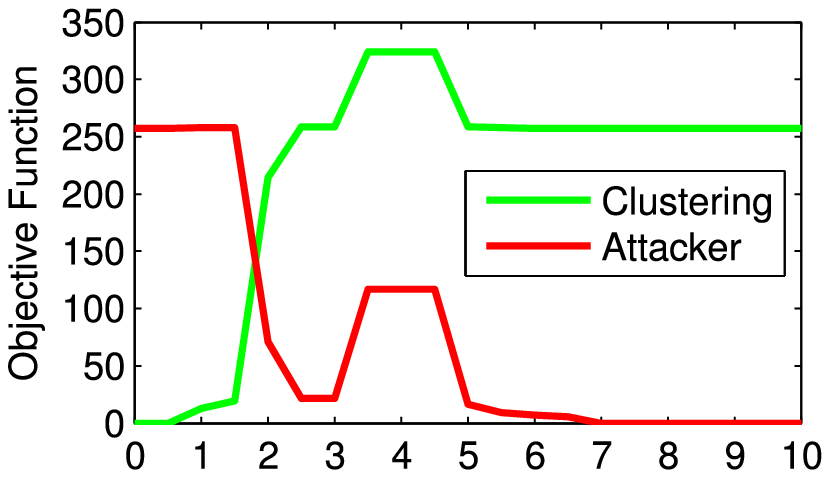} \\
\includegraphics[width=0.32\textwidth]{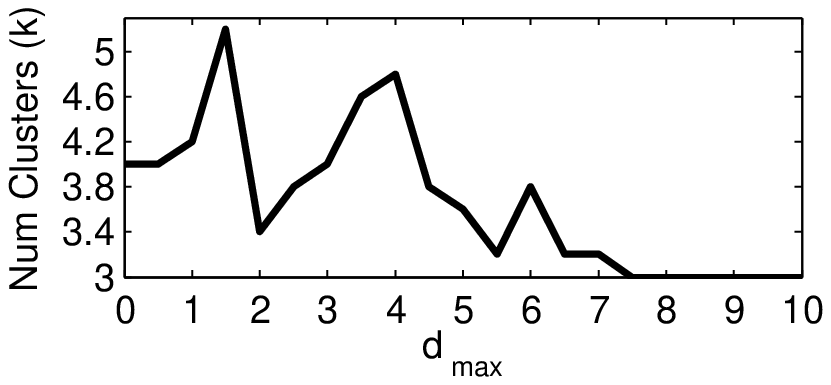}
\vspace{-10pt}
\caption{Results for the obfuscation attack averaged over five runs on the Digit dataset. The top plots shows the variation of the objective function for the attacker $d_{c}( \set C^{t},f(\set D \cup \set A^{\prime}))$ and for the clustering algorithm $d_{c}( f(\set D \cup \set A),f(\set D \cup \set A^{\prime}))$ as the maximum amount of modifications $d_{\rm max}$ to the initial attack samples $\set A$ increases. The bottom plot reports the corresponding average number of selected clusters.}
\label{fig:results-obfuscation}
\end{figure}
\begin{figure}[t]
\centering
\includegraphics[width=\linewidth]{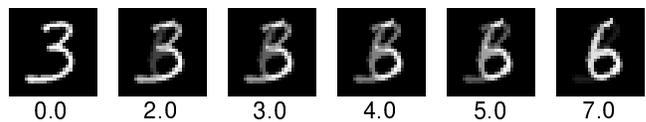} \\
\vspace{-10pt}
\caption{An example of how a digit `3' is gradually manipulated to resemble the closest `6', for different values of $d_{\rm max}$.}
\label{fig:results_obfuscation_chars}
\vspace{-10pt}
\end{figure}

{
  \setlength{\parskip}{3pt}
  \setlength{\parsep}{3pt}

On the other hand, as explained in Sect.~\ref{sect:obfuscation-attacks}, the clustering algorithm attempts to keep the attack points corresponding to the digit `3' into a well-separated cluster from the remaining digits, \ie, it selects the number of clusters that minimizes $d_{c}( \set C^{\star},f(\set D \cup \set A^{\prime}))$, which can thus be regarded as the objective function for the clustering algorithm. In this case, $\set C^{\star}$ is the  clustering obtained on the initial data and the non-manipulated attack samples, \ie, $C^{\star}=f(\set D \cup \set A)$.

The results for the above discussed obfuscation attack are given in Fig.~\ref{fig:results-obfuscation}, where we report the values of the objective function for the attacker and for the clustering algorithm, as well as the number of selected clusters, as a function of the maximum amount of allowed modifications to the attack samples, given in terms of the maximum Euclidean distance $d_{\rm max}$ (see Eq.~\ref{eq:d_s}).
The results clearly show that the objective function of the attacker tends to decrease, while that of the clustering algorithm generally increases. The reason is that, initially, the clustering algorithm correctly separates the four clusters associated to the four distinct digits, whereas as $d_{\rm max}$ increases, the attack digits `3' are more and more altered to resemble the closest `6's, and are then gradually merged to their cluster.
The number of clusters does not decrease immediately to $3$ as one would expect since, while manipulating the attack samples, their cluster is fragmented into smaller ones (typically, two or three clusters). The reason is that, to remain as close as possible to the ideal $\set C^{\star}$, the clustering algorithm avoids some of the `3's to immediately join the cluster of `6's by fragmenting the cluster of `3's. 

When $d_{\rm max}$ takes on values approximately in $[3,4]$, the clustering algorithm creates only three clusters, corresponding effectively to the attacker's goal $\set C^{t}$ (this is witnessed by the fact that the averaged attacker's objective is almost zero). Surprisingly, though, as soon as $d_{\rm max}$ becomes greater than 4, the number of clusters raises again to 4, and some of the attack samples are again separated from the cluster of `6's, worsening the adversary's objective. This is due to the fact that, when $d_{\rm max} \approx 3$ or $4$, some of the attack points work as \emph{bridges} and successfully connect the remaining `3's to the cluster of `6's, whereas when these points are further shifted towards the cluster of `6's, the algorithm can successfully split the two clusters again.
Based on this observation, a \emph{smarter} attacker may even manipulate only a very small subset of her attack samples to create proper \emph{bridges} and connect the remaining non-manipulated samples to the desired cluster. We however left a quantitatively investigation of this approach to future work.

In Fig.~\ref{fig:results_obfuscation_chars} we finally report an example of how a digit `3' is manipulated by our attack to be hidden in the cluster associated to the digit `6'.
It is worth noting how, when $d_{\rm max} \in [2, 4]$, the original attack sample still significantly resembles the initial `3': this shows that the adversary's goal can be achieved without altering too much the initial attack samples, which is clearly a strong \emph{desideratum} for the attacker in adversarial settings.
}

\section{Conclusions and future work}
\label{sect:conclusions}

In this paper, we addressed the problem of evaluating the security of clustering algorithms in adversarial settings, by providing a framework for simulating potential attack scenarios. We devised two attacks that can significantly compromise availability and integrity of the targeted system. We demonstrated with real-world experiments that single-linkage clustering may be significantly vulnerable to deliberate attacks, either when the adversary can only control a very small fraction of the input data, or when she slightly manipulates her attack samples.
This shows that attacking clustering algorithms with tailored strategies can significantly alter their output to meet the adversary's goal. 

Admittedly, one of the causes of the vulnerability of single-linkage resides in its inter-cluster distance, which solely depends on the closest points between clusters, and thus allowed for an efficient constructing of bridges. It is reasonable to assume that algorithms based on computing averages (\eg, k-means) or density estimation might be more robust to poisoning, although not necessarily robust to obfuscation attacks. However, the results of our empirical evaluation can not be directly generalized to different algorithms, and more investigation should thus be carried out in this respect.

In general, finding the optimal attack strategy given an arbitrary clustering algorithm is a hard problem and we have to rely on heuristic algorithms in order to carry out our analysis. 
For the sake of efficiency, these heuristics should be heavily dependent on the targeted clustering algorithm, as in our case. However, it would be interesting to exploit more general approaches that ideally treat the clustering algorithm as a black box and find a solution by performing a stochastic search on the solution space (\eg{} by simulated annealing), or an educated exhaustive search (\eg{} by using branch-and-bound techniques).

{
  \setlength{\parskip}{3pt}
  \setlength{\parsep}{3pt}

In this work we did not address the problem of \emph{countering attacks}
by designing more \emph{secure} clustering algorithms. We only assumed that the clustering algorithm can select a different number of clusters (optimal according to its goal) after each attack iteration.
More generally, though, one can design a clustering algorithm that explicitly takes into account the adversary's presence, and her optimal attack strategy, \eg, by modeling clustering in adversarial settings as a game between the clustering algorithm and the attacker. This has been done in the case of supervised learning, to improve the security of learning algorithms against evasion attempts \cite{brueckner12}, and similarly, in the regression setting \cite{grosshans13}. Other approaches may more directly encode explicit assumptions on how the data distribution changes under attack, similarly to \cite{biggio11-smc}. We left this investigation to future work.

Another possible future extension of our work would be to consider a more realistic setting in which the attacker has limited knowledge of the attacked system.
To this end, the upper bound on the performance degradation incurred under attack provided by our worst-case analysis may be exploited to evaluate the effectiveness of attacks devised under limited knowledge (\ie, how close they can get to the worst case).

One limitation of our approach may be the so-called \emph{inverse feature-mapping} problem~\cite{huang11,biggio13-tkde}, \ie, the problem of finding a real attack sample corresponding to a desired feature vector (as the ones suggested by our attack strategies). In the reported experiments, this was not a significant problem since modifications to the given feature values could be directly mapped to manipulations on the \emph{real} attack samples. Although inverting the feature mapping may be a cumbersome task for more complicated feature representations, this remains a common problem of optimal attacks in adversarial learning, and it has to be addressed in an application-specific manner, depending on the given feature space.

As a further future development, we plan to establish a link between the evaluation of the security of clustering algorithms and the problem of determining the \emph{stability} of a clustering, which has been already addressed in the literature and used as a device for model selection (see, \eg, \cite{Lux10}). Indeed, stable clusterings can be regarded as secure under specific non-targeted attacks like, \eg{}, perturbation of points with Gaussian noise.

Understanding robustness of clustering algorithms against carefully targeted attacks under a more theoretical perspective (\eg, by devising theoretical bounds that evaluate the impact of single attack points on the clustering output) may also be a promising research direction. Some results from clustering stability may be also exploited to this end.

}

\section{Acknowledgments}
This work has been partly supported by the Regional Administration of Sardinia (RAS), Italy, within the projects  ``Security of pattern recognition systems in future internet'' (CRP-18293), and ``Advanced and secure sharing of multimedia data over social networks in the future Internet'' (CRP-17555). Both projects are funded within the framework of the regional law \emph{L.R. 7/2007, Bando 2009}. 
The opinions, findings and conclusions expressed in this paper are solely those of the authors and do not necessarily reflect the opinions of any sponsor.

\end{document}